\documentclass[sn-basic]{sn-jnl}

\jyear{2021}%

\theoremstyle{thmstyleone}%
\theoremstyle{thmstyletwo}%

\theoremstyle{thmstylethree}%

\usepackage{booktabs}
\usepackage{soul}
\usepackage{amsmath}
\usepackage{amssymb}
\usepackage{makecell}
\usepackage{bbding}
\usepackage{pifont}
\usepackage{bigstrut}
\usepackage{color}
\usepackage{multirow}
\usepackage{algorithm}
\usepackage{algorithmicx}
\usepackage{algpseudocode}

\begin{document}

\title[Learning Discriminative Visual-Text Representation for Polyp Re-ID]{Learning Discriminative Visual-Text Representation for Polyp Re-Identification}


\author*[1]{\fnm{Suncheng} \sur{Xiang}}\email{xiangsuncheng17@sjtu.edu.cn}

\author[2]{\fnm{Cang} \sur{Liu}}\email{liucang@alumni.nudt.edu.cn}

\author[1]{\fnm{Sijia} \sur{Du}}\email{Scarlett\_Du@sjtu.edu.cn}

\author[1]{\fnm{Dahong} \sur{Qian}}\email{dahong.qian@sjtu.edu.cn}


\affil[1]{\orgdiv{School of Biomedical Engineering}, \orgname{Shanghai Jiao Tong University}, \orgaddress{\city{Shanghai}, \postcode{200240}, \country{China}}}

\affil[2]{\orgdiv{National Key Laboratory of Electromagnetic Energy}, \orgname{Naval University of Engineering}, \orgaddress{\city{Wuhan}, \postcode{430033}, \country{China}}}


%


\abstract{Colonoscopic Polyp Re-Identification aims to match a specific polyp in a large gallery with different cameras and views, which plays a key role for the prevention and treatment of colorectal cancer in the computer-aided diagnosis. However, traditional methods mainly focus on the visual representation learning, while neglect to explore the potential of semantic features during training, which may easily leads to poor generalization capability when adapted the pretrained model into the new scenarios. To relieve this dilemma, we propose a simple but effective training method named \textbf{VT-ReID}, which can remarkably enrich the representation of polyp videos with the interchange of high-level semantic information.
Moreover, we elaborately design a novel clustering mechanism to introduce prior knowledge from textual data, which leverages contrastive learning to promote better separation from abundant unlabeled text data.
To the best of our knowledge, this is the first attempt to employ the visual-text feature with clustering mechanism for the colonoscopic polyp re-identification. Empirical results show that our method significantly outperforms current state-of-the art methods with a clear margin.}





\maketitle

\section{Introduction}
\label{sec1}
Colonoscopic polyp re-identification (Polyp ReID) aims to match a specific polyp in a large gallery with different cameras and locations, which has been studied intensively due to its practical importance in the prevention and treatment of colorectal cancer in the computer-aided diagnosis. With the development of deep convolution neural networks and the availability of video re-identification dataset, video retrieval methods have achieved remarkable performance in a supervised manner~\citep{huang2021cross,feng2019spatio}, where a model is trained and tested on different splits of the same dataset. However, in practice, manually labelling a large diversity of pairwise polyp area data is time-consuming and labor-intensive when directly deploying polyp ReID system to new hospital scenarios~\citep{chen2023colo}.
As a special case of image retrieval, the polyp re-identification has also drawn increasing research interests due to its vast market~\citep{chen2023colo}. Nevertheless, polyp is confronted with more challenges, compared to the conventional person re-identification~\citep{xiang2021less,xiang2023deep}. Firstly, the backgrounds of the videos are more complex and diverse. Secondly, brightness / viewpoint variation in the practical setting can be observed rather different videos. Thirdly, the involved videos noises often come from multi heterogeneous modalities, ranging from image, sound and text information.


To address these problem, previous methods such as \cite{chen2023colo} propose a self-supervised contrastive representation learning scheme named Colo-SCRL to learn spatial representations from colonoscopic video dataset. \cite{xiang2020unsupervised} design an unsupervised ReID method via domain adaptation on the basis of  a diverse synthetic dataset. Some other researchers~\citep{chen2020fine,dong2016word2visualvec,li2020adversarial} struggle to find a representative video frame, and then feed it into the image-text model for video-text retrieval. Besides, \cite{xu2015jointly} propose a unified framework that jointly models video and the corresponding text sentences, which can leverage deep neural networks to capture essential semantic information from videos. Although these approaches can achieve promising progress, their performance deeply relies on the scale/number of polyp training dataset. Importantly, other rich information in the videos effective for video-text retrieval is ignored.

\begin{figure*}[!t]
\centering{\includegraphics[width=\linewidth]{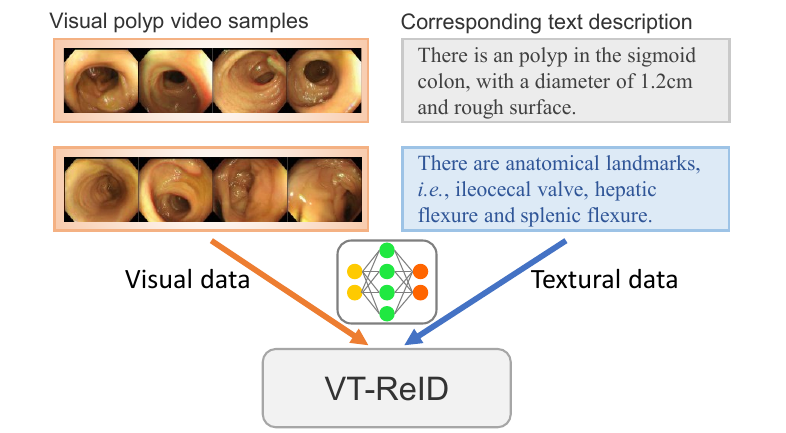}}
\caption{Corresponding text description of visual polyp samples, which is excerpted from the colonoscopy examination and treatment report form. The main goal of this work is to learn a robust polyp re-identification model on the basis of visual-text representation.}
\label{fig1}
\end{figure*}

In order to alleviate the above-mentioned problems in terms of data scale and learning paradigm, we adopt the concept of multimodal learning for more robust visual-semantic embedding learning for polyp ReID event. On the basis of it, a dynamic learning strategy called \textbf{DCM} is introduced to further enhance the visual-text feature fusion in multimodal polyp ReID task, which can jointly work on unimodal visual or textual modality and significantly boost the performance of polyp ReID task. As illustrated in Fig.~\ref{fig1}, our model is trained on a set of source domain based on the unpaired images/texts or image-text pairs, and should generalize to any new unseen datasets for effective Re-ID without any model updating.
During the training process, we adopt the instance-contrastive loss as optimization function in our algorithm,
which can pull together samples augmented from the same instance in the original dataset while pushing apart those from different ones. Essentially, instance-contrastive learning disperses different instances apart while implicitly bringing similar instances together to some extent.
To the best of our knowledge, this is the first research effort to exploit the potential of multimodal feature on the basis of polyp dataset to address the polyp ReID task. We hope this method could advance
research towards medical understanding.

Compared with traditional polyp re-identification approach, our VT-ReID method is different from them in terms of two perspective: \textbf{data input} and \textbf{model structure}:
(1) First, previous methods always require large-scale datasets for model training to obtain the competitive retrieval performance, while our VT-ReID model works on a wide range of retrieval tasks in a scenario with few samples, which allows our model to be more flexible and adaptable in real-world scenario;
(2) Our \textbf{VT-ReID} method adopts the two-stream branches for visual-texture representation, which considers the interchange of high-level semantic information among multiple features and successfully retrieves the correct polyp video when given an complex query. However, previous approaches fail to take advantage of multimodal feature since most of them are based on the single modal structural design.

As a consequence, the major contributions of our work can be summarized into three-fold:
\begin{itemize}
 \item[$\bullet$] We propose a simple but effective training method named \textbf{VT-ReID}, which helps our model to learn general visual-text representation based on the multimodal feature.

 \item[$\bullet$] Based on it, a dynamic clustering mechanism called \textbf{DCM} is introduced to further enhance the clustering performance of text data in an unsupervised manner.

 \item[$\bullet$] Comprehensive experiments show that our \textbf{VT-ReID} method matches or exceeds the performance of existing methods with a clear margin, which reveals the applicability of colonoscopic polyp re-identification task with new insights.
\end{itemize}

In the rest of the paper, we first review some related works of colonoscopic polyp re-identification methods and previous deep learning-based methods in Section~\ref{sec2}. Then in Section~\ref{sec3}, we give more details about the learning procedure of the proposed Colo-ReID method. Extensive evaluations compared with state-of-the-art methods and comprehensive analyses of the proposed approach are elaborated in Section~\ref{sec4}. Conclusion and Future Works are given in Section~\ref{sec5}.

\section{Related Works}
\label{sec2}
In this section, we have a brief review on the related works of traditional re-identification methods. The mainstream idea of the existing methods is to learn a discriminative and robust model for downstream task. There methods can be roughly divided into hand-crafted based approaches and deep learning based approaches.

\subsection{Hand-crafted based Approaches}

Traditional research works~\citep{prosser2010person,zhao2014learning,Zhao2013PersonRB} related to hand-crafted systems for image retrieval task aim to design or learn discriminative representation or pedestrian features. For example, \cite{prosser2010person} propose a reformulation of the person re-identification problem as a learning to rank problem.
\cite{Zhao2013PersonRB} exploit the pairwise salience distribution relationship between pedestrian images, and solve the person
re-identification problem by proposing a salience matching strategy. Besides directly using mid-level color and texture features, some methods~\citep{zhao2014learning,xiang2020multi} also explore different discriminative abilities of local patches for better discriminative power and generalization ability.
Unfortunately, these hand-crafted feature based approaches always fail to produce competitive results on large-scale datasets. The main reason is that these early works are mostly based on heuristic design, and thus they could not learn optimal discriminative features on current large-scale datasets.

\subsection{Deep learning based Approaches}

Recently, there has been a significant research interest in the design of deep learning based approaches for video retrieval~\citep{shao2021temporal,xiang2022learning,ma2020vlanet,xiang2020multi}. For example, \cite{shao2021temporal} propose temporal context aggregation which incorporates long-range temporal information for content-based video retrieval. \cite{xiang2020multi} propose a feature fusion strategy based on traditional convolutional neural network for pedestrian retrieval.
\cite{ma2020vlanet} explore a method for performing video moment retrieval in a weakly-supervised manner. \cite{lin2020weakly} propose a semantic completion network including the proposal generation module to score all candidate proposals in a single pass. Besides, \cite{kordopatis2022dns} propose a video retrieval framework based on knowledge distillation that addresses the problem of performance-efficiency trade-off.
As for the self-supervised learning, \cite{wu2018unsupervised} present an unsupervised feature learning approach called Instance-wise Contrastive Learning (Instance-CL)~\citep{wu2018unsupervised} by maximizing distinction between instances via a non-parametric softmax formulation.
In addition, \cite{chen2023colo} propose a self-supervised contrastive representation learning scheme named Colo-SCRL to learn spatial representations from colonoscopic video dataset.
However, all above approaches divide the learning process into multiple stages, each requiring independent optimization. On the other hand, despite the tremendous successes achieved by deep learning-based approach, they have been left largely unexplored for short text clustering.


To address the challenges mentioned above, we propose a feature learning strategy combining global and local information in different modalities (named "VT-ReID" for short),
which helps our model learn robust and discriminative visual-text representation with multiple granularities. Based on it, a dynamic clustering mechanism called \textbf{DCM} is introduced to promote the clustering results of text annotation in an unsupervised manner, which can significantly improve the performance of Colo-ReID method on colonoscopic polyp ReID task. Specifically, our model adopts multiple branches to learn visual and text representation simultaneously, where first branch employs the MoCo mechanism to perform unsupervised visual representation learning, the second branch use contrastive learning to perform clustering, which can significantly capture the common features in the colonoscopy video and improve the stability of the model. To the best of our knowledge, this is the first attempt to employ the contrastive learning paradigm on clustering for colonoscopic polyp ReID task. We hope this research work could advance research towards medical understanding, especially for the medical identification and retrieval task.

\section{Our Proposed Approach}
\label{sec3}

\subsection{Preliminary}
We begin with a formal description of common polpy ReID problem. Assuming that we are given with a source domain $\mathcal{D}$, which contains its own label space $\mathcal{D}=\left\{\left(\boldsymbol{x}_i, y_i\right)\right\}_{i=1}^{N}$ in terms of visual and textual feature, where $N$ is the number of images in the source domain $\mathcal{D}$. Each sample $\boldsymbol{x}_i \in \mathcal{X}$ is associated with an identity label $y_i \in \mathcal{Y}=\left\{1,2, \ldots, M\right\}$, where $M$ is the number of identities in the source domain $D$, along with the corresponding text description $\boldsymbol{x}_t$ of colonoscopic polyp dataset.
During the training stage, we need to train a robust polyp Re-ID model using the aggregated image-label pairs of source domain. In the testing stage, we perform a retrieval task on unseen target domains without additional model updating.

\subsection{The Proposed VT-ReID Framework}

\begin{figure*}[!t]
\centering{\includegraphics[width=\linewidth]{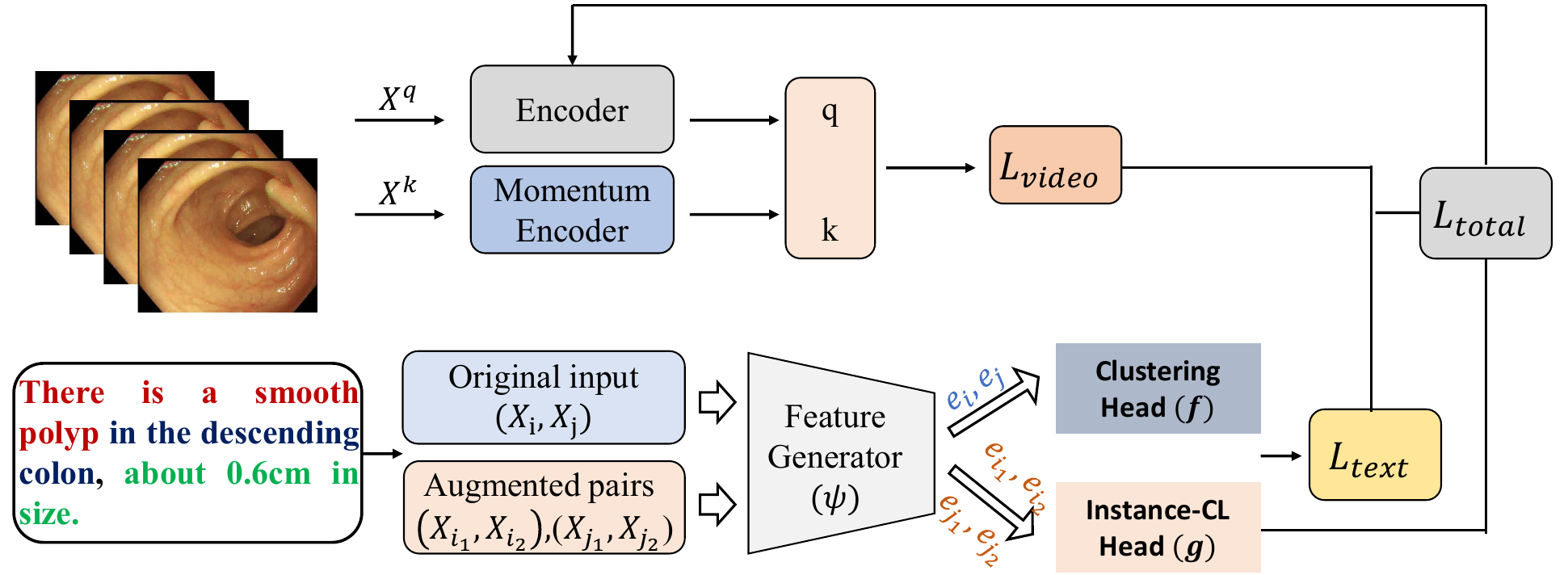}}
\caption{The proposed VT-ReID Network architecture. The visual feature backbone is composed with vision transformer, while texture feature backbone is composed with deep neural network, which can map the input data to the representation space. Finally, our method will learn visual and textual feature in an end-to-end manner for more robust and discriminative representation.}
\label{fig2}
\end{figure*}

We aim at developing a joint model that leverages the beneficial properties of instance-wise contrastive learning to improve the performance of colonoscopic polyp retrieval task.
As illustrated in Fig.~\ref{fig2}, our model adopts the vision transformer~\citep{dosovitskiy2020image} to perform video representation extraction, and employs the supporting clustering with contrastive learning~\citep{zhang2021supporting} for text representation extraction respectively. Then, we combine the visual and textural feature for the downstream image retrieval task during the inference period.
The training process of \textbf{VT-ReID } can be described in two aspects. First, during the visual feature extraction stage, we maintain the dictionary as a \textit{queue} of data samples: the encoded representations of the current mini-batch are enqueued, and the oldest are dequeued. Second, during the texture feature extraction stage, the contrastive learning paradigm is employed to further enhance the performance of clustering, which can encourage learning from high confidence cluster assignments and simultaneously combating the bias caused by imbalanced clusters.

\textbf{Visual Representation Learning.}
Contrastive learning~\citep{hadsell2006dimensionality}, and its recent developments, can be thought of as training an encoder for \textit{dictionary look-up} task, as described next.
Inspired by the previous work MoCo~\citep{he2020momentum}, we adopt the Momentum Contrast for unsupervised visual representation learning. To be more specific, we maintain the dictionary as a \textit{queue} of data samples: the encoded representations of the current mini-batch are enqueued, and the oldest are dequeued. The queue decouples the dictionary size from the mini-batch size, allowing it to be large. To be more specific, we adopt a queue to store and sample negative samples, the queue stores multiple feature vectors of recently used batches for training. The queue can be continuously updated, thus reducing storage consumption.

Consider an encoded query $q$ and a set of encoded samples $\left\{k_0, k_1, k_2, \ldots\right\}$ that are the keys of a dictionary. Assume that there is a single key (denoted as $k_{+}$) in the dictionary that $q$ matches. A contrastive loss~\citep{hadsell2006dimensionality} is a function whose value is low when $q$ is similar to its positive key $k_{+}$ and dissimilar to all other keys (considered negative keys for $q$). With similarity measured by dot product, a form of a contrastive loss function, called InfoNCE~\citep{oord2018representation}, is considered in this paper:

\begin{equation}
\mathcal{L}_{video}=-\log \frac{\exp \left(q \cdot k_{+} / \tau\right)}{\sum_{i=0}^K \exp \left(q \cdot k_i / \tau\right)}
\label{eq1}
\end{equation}
where $\tau$ is a temperature hyper-parameter of $\mathcal{L}_{video}$, which controls the distribution of logistic function. The sum $\sum_{i=0}^K$ is over one positive and $K$ negative samples.


\textbf{Text Representation Learning.}
In essence, most of the previous cross-modal retrieval approaches try to learn a joint embedding space to measure the cross-modal similarities, while paying little attention to the representation of text modality. To address this problem, we propose a dynamic contrastive learning-based clustering mechanism to further enhance the clustering performance of text data in an unsupervised manner. As depicted in Fig.~\ref{fig2} (\textbf{lower part}), our clustering algorithm contains three parts: Feature Extraction Layer, Clustering Head and Instance-CL Head, it is worth mentioning that the feature extraction layer can map the input into a high-dimension vector space.
To be more specific, we employ some data augmentation method to enhance the sample data for obtaining an auxiliary dataset, and then perform optimizing based on this dataset. During the optimization process, we also employ contrastive loss to make the enhanced samples from the same instance approach each other in the representation space, while the enhanced samples from different instances stay away from each other in the representation space, this constraint can lead to a better separation of different clusters, even with smaller distances between samples within the cluster. Consequently, we simplify the learning process to end-to-end training with the target distribution being updated per iteration.

\begin{equation}
\mathcal{L}_{text}=\sum_{j=1}^M \ell_j^C / M+\eta * \sum_{i=1}^{2 M} \ell_i^I / 2 M
\label{eq2}
\end{equation}
where $M$ represents the batchsize of mini-batch during training, $\ell_j^C $ and $\ell_i^I$ denote the KL divergence of clustering and Instance-Contrastive Learning respectively. And $\eta$ balances between the contrastive loss and the clustering loss of supporting clustering with contrastive learning~\citep{zhang2021supporting}.

In this work, we propose to jointly learn the video information loss calculated by MoCo~\citep{he2020momentum} and the text loss calculated by supporting clustering, more detailed training procedure is illustrated in Algorithm~\ref{alg1}. Specifically, there are labeled polyp videos, and their corresponding diagnostic report. Finally, the overall objective loss function in a training batch is expressed as:

\begin{equation}
\mathcal{L}_{total}=\mathcal{L}_{video}+\mathcal{L}_{text}
\label{eq3}
\end{equation}

\begin{algorithm}[!t]
	\renewcommand{\algorithmicrequire}{\textbf{Input:}}
	\renewcommand{\algorithmicensure}{\textbf{Output:}}
	\caption{The training procedure of our proposed VT-ReID method}
	\label{alg1}
	\begin{algorithmic}[1]
		\Require \\
        Image sample  $\boldsymbol{x}_i$ of colonoscopic dataset for training, \\
        Text sample  $\boldsymbol{x}_t$ of colonoscopic dataset for training, \\
        Iteration $n$ for model updating;
		\Ensure
		Polyp Re-ID model $f(x)$;
		\State Initialize: $iter$=1;
        \State $\rhd$ \textcolor[rgb]{0.11,0.21,0.65}{Visual feature extraction:}
        \State Extract visual feature $\boldsymbol{f}_i$ from image sample  $\boldsymbol{x}_i$;
        \State Perform momentum contrast with $\boldsymbol{f}_i$ for visual representation;
        \State $\rhd$ \textcolor[rgb]{0.11,0.21,0.65}{Text feature extraction:}
        \State Extract text feature $\boldsymbol{f}_t$ from text sample  $\boldsymbol{x}_t$;
        \State Perform contrastive learning-based clustering with $\boldsymbol{f}_t$ for text representation;
        \If {iter $\leq$ n}
        \State Optimize model $f(x)$ with video loss $\mathcal{L}_{video}$ and text loss $\mathcal{L}_{text}$;
        \State \textit{iter} $\leftarrow$ \textit{iter} + 1;
        \EndIf  \\
        \Return Optimized model $f(x)$;
	\end{algorithmic}
\end{algorithm}


To sum up, the proposed VT-ReID network is trained to minimize the two loss functions jointly. To figure out which objective function contributes more, we train the identification model with different loss functions separately, following the learning rate setting in Section~\ref{sec4.2}. First, we train the models with text loss function. Then, video loss function is employed with text loss to train the network jointly until two objectives both reaching convergence. The experiment details can be accessed in Section~\ref{sec4.3}, our polyp ReID model with two kinds of objective loss outperforms the one trained individually. This result has been confirmed on the backbone of vision transformer network structures.

\section{Experimental Results}
\label{sec4}
\subsection{Datasets}
\label{sec4.1}
In this paper, we conduct experiments on several large-scale
public datasets, which include Colo-Pair~\citep{chen2023colo}, Market-1501~\citep{zheng2015scalable}, DukeMTMC-reID~\citep{ristani2016performance,zheng2017unlabeled} and CUHK03 dataset~\citep{li2014deepreid}.
The evaluation protocols are also elaborated before showing our experiment results.

\textbf{Market-1501}~\citep{zheng2015scalable} consists of 32,668 person images of 1,501 identities observed under 6 different camera views. The dataset is split into 12,936 training images of 751 identities and 19,732 testing images of the remaining 750 identities. Both training and testing images are detected using a DPM detector~\citep{felzenszwalb2009object}.

\textbf{DukeMTMC-reID}~\citep{ristani2016performance,zheng2017unlabeled} is collected in the winter of Duke University from 8 different cameras, which contains 16,522 images of 702 identities for training, and the remaining images of 702 identities for testing, including 2,228 images as query and 17,661 images as gallery, the bounding-box of this dataset are all manually annotated.

\textbf{CUHK03}~\citep{li2014deepreid} contains 14.096 images which are collected by the Chinese University of Hong Kong, the images are taken from only 2 cameras. In fact, CUHK03 dataset provides two types of data with manual labels and DPM detection bounding boxes. In this work, we conduct experiments on the labeled bounding boxes (CUHK03 (Labeled)).

\textbf{Colo-Pair}~\citep{chen2023colo} is the first collection of complete paired colonoscopy sequences acquired during regular medical practice in total 9.6 hours of paired videos with full procedures, \textit{i.e.}, slow and careful screening explorations, which contains 60 videos from 30 patients, with 62 query video clips and the corresponding polyp clips from the second screening manually annotated as positive retrieval clips.

\textbf{Evaluation Metric}
In our experiments, we follow the standard evaluation protocol~\citep{zheng2015scalable} used in Re-ID task , and adopt mean Average Precision (mAP) and Cumulative Matching Characteristics (CMC) at Rank-1, Rank-5 and Rank-10 for performance evaluation on downstream Re-ID task.
These metrics can evaluate the performance of the colon polyp recognition model. By using different evaluation metrics, we can comprehensively measure the model's effectiveness and compare its performance with others.

\subsection{Implementation Details}
\label{sec4.2}
In our experiment, Vision-Transformer (ViT)~\citep{dosovitskiy2020image} is regarded as the backbone with no bells and whistles during in visual feature extraction. Following the training procedure in~\cite{Tong2022VideoMAEMA}, we adopt the common methods such as random flipping and random cropping for data augmentation, and employ the Adam optimizer with a weight decay co-efficient of $1 \times 10^{-5}$ and $1 \times 10^{-7}$ for parameter optimization. During the training, we adopt the video loss and text loss functions to train the model for 180 iterations, where temperature hyper-parameter in Eq.~\ref{eq1} is empirically set as 0.07, and the hyper-parameter $\eta$ for the loss function in Eq.~\ref{eq2} is empirically set to 10 for simplicity. In addition, the batch size $M$ for training is set to 64.
All the experiments are performed on PyTorch~\citep{paszke2019pytorch} with one Nvidia GeForce RTX 2080Ti GPU on a server equipped with a Intel Xeon Gold 6130T CPU.

\subsection{Ablation study}
\label{sec4.3}
In order to prove the effectiveness of individual technical contributions, we perform the following ablation studies in terms of proposed VT-ReID method and dynamic clustering mechanism.

\textbf{The effectiveness of VT-ReID method.}
To validate the effectiveness of different components, we conduct several ablation experiments to validate the effectiveness of our proposed multimodal method. For example, when adding the text loss to the VT-ReID method, as illustrated in Table~\ref{tab1}, our text loss is consistently improved in all settings. Specially, the mAP accuracy
 ``Baseline+$\mathcal{L}_{text}$" on Colo-Pair is increased from 25.9\% to 27.3\% on polyp ReID event. We can also get the Rank-1 accuracy of 21.7\% when applying the $\mathcal{L}_{video}$ in our VT-ReID method. Especially, when applying the text loss $\mathcal{L}_{text}$ and video loss $\mathcal{L}_{video}$ together on Colo-Pair dataset, our method VT-ReID ($\mathcal{L}_{text}$+$\mathcal{L}_{video}$) can get a mAP accuracy of 37.9\% on Colo-Pair dataset, which surpasses the baseline by \textbf{+12.0\%} and \textbf{+5.9\%} in terms of mAP and Rank-1 accuracy respectively. During the ablation study, significant improvement is also observed in other setting, so the use of visual-text representation has a prominent contribution to the performance of our VT-ReID model.

\begin{table}[!t]
  \centering
  \caption{Ablation study on the text loss and video loss.}
  \setlength{\tabcolsep}{2.7mm}{
    \begin{tabular}{lccccc}
    \toprule
    Methods  & Text data  & Image data   & mAP$\uparrow$   & Rank-1$\uparrow$ & Rank-5$\uparrow$ \\
    \midrule
    Baseline  & $\times$  & $\times$  & 25.9  & 17.5  & 37.1 \\
    \textbf{w/} text loss $\mathcal{L}_{text}$   & $\checkmark$  & $\times$  & 27.3  & 17.4  & 38.4 \\
     \textbf{w/} video loss $\mathcal{L}_{video}$  & $\times$  & $\checkmark$   & 31.6  & 21.7  & 41.2 \\
     \textbf{w/} total loss $\mathcal{L}_{total}$  & $\checkmark$  & $\checkmark$  & 37.9  & 23.4  & 44.5 \\
    \bottomrule
    \end{tabular}}%
  \label{tab1}%
\end{table}%


\textbf{The effectiveness of dynamic clustering mechanism.}
In this section, we evaluate the effectiveness of contrastive-based clustering strategy. As illustrated in Table~\ref{tab1}, the results show that mAP accuracy drops significantly from 37.9\% to 31.6\% without adopting the support clustering mechanism.
Additionally, similar drops can also be observed no matter which evaluation metric is employed in our multimodal visual-text feature learning framework (\textit{e.g.} \textcolor[rgb]{1.00,0.39,0.09}{\textbf{23.4\%}} vs. \textcolor[rgb]{0.20,0.40,0.80}{\textbf{21.7\%}} in Rank-1 accuracy). The effectiveness of our the dynamic clustering mechanism can be largely attributed to that it enhances the discriminative capability of all multimodal networks during visual-text representation learning, which is vital for polyp re-identification in cross-domain where the target supervision is not available.

\subsection{Comparison to the state-of-the-art models}
In this section, we mainly validate the effectiveness of our method on colonoscopic polyp re-identification and person re-identification tasks respectively.

\textbf{Colonoscopic Polyp Re-Identification.}
To prove the effectiveness of our method, we compare our method with the state-of-the-art object re-identification models including: ViSiL~\citep{kordopatis2019visil}, CoCLR~\citep{han2020self}, TCA~\citep{shao2021temporal}, ViT~\citep{caron2021emerging}, CVRL~\citep{qian2021spatiotemporal}, CgS$^c$~\citep{kordopatis2022dns}, FgAttS$^f_A$~\citep{kordopatis2022dns}, FgBinS$^f_B$~\citep{kordopatis2022dns} and Colo-SCRL~\citep{chen2023colo}, respectively.

Table~\ref{tab2} shows the clear performance superiority of VT-ReID over all state-of-the-arts with significant Rank-1 and mAP advantages. Specifically, our VT-ReID outperforms the second best model Colo-SCRL~\citep{chen2023colo} by \textbf{+6.4\%} (37.9-31.5) and \textbf{+0.8\%} (23.4-22.6) in terms of mAP and Rank-1 accuracy, respectively. Compared to the transformer-only (soft attention) network ViT~\citep{caron2021emerging}, our model improves the Rank-1 by \textbf{+13.7\%} (23.4-9.7). This indicates the superiority of our visual-text representation learning mechanism. VT-ReID also surpasses recent knowledge distillation-based methods, such as CgS$^c$, FgAttS$^f_A$ and FgBinS$^f_B$~\citep{kordopatis2022dns}, boosting the Rank-1 by \textbf{+15.3\%}, \textbf{+13.7\%} and  \textbf{+13.7\%}, mAP by \textbf{+16.5\%}, \textbf{+14.3\%} and \textbf{+16.7\%}, respectively. These analyses validate the significant advantage of our visual-text representation learning over existing methods replying on either attention or knowledge distillation-based methods at a single level.

\begin{table}[!t]
\centering
\caption{Performance comparison with other SOTA  methods (measured by \%) on the standard Colo-Pair Dataset. \textbf{Bold} indicates the best and \underline{underline} the second best.}
\setlength{\tabcolsep}{1.52mm}{
\begin{tabular}{lccccc}
\toprule
\multirow{2}{*}{Method} & \multirow{2}{*}{Venue} & \multicolumn{4}{c}{Video Retrieval $\uparrow$} \\
\cmidrule{3-6}  &  & mAP & Rank-1 & Rank-5 & Rank-10  \\
\midrule
ViSiL~\citep{kordopatis2019visil} & ICCV 19 & 24.9 & 14.5 & 30.6 & 51.6  \\
CoCLR~\citep{han2020self} & NIPS 20 & 16.3 & 6.5 & 22.6 & 33.9  \\
TCA~\citep{shao2021temporal} & WCAV 21 & 27.8 & 16.1 & 35.5 & 53.2  \\
ViT~\citep{caron2021emerging} & CVPR 21 & 20.4 & 9.7 & 30.6 & 43.5  \\
CVRL~\citep{qian2021spatiotemporal} & CVPR 21 & 23.6  & 11.3 & 32.3 & 53.2  \\
CgS$^c$~\citep{kordopatis2022dns} & IJCV 22 & 21.4  & 8.1 & 35.5 & 45.2  \\
FgAttS$^f_A$~\citep{kordopatis2022dns} & IJCV 22 & 23.6  & 9.7 & 40.3 & 50.0  \\
FgBinS$^f_B$~\citep{kordopatis2022dns} & IJCV 22 & 21.2  & 9.7 & 32.3 & 48.4  \\
Colo-SCRL~\citep{chen2023colo} & ICME 23 & \underline{31.5}  & \underline{22.6} & \underline{41.9} & \underline{58.1}  \\
\midrule
\textbf{VT-ReID} & \textbf{Ours} & \textbf{37.9} & \textbf{23.4} & \textbf{44.5} & \textbf{60.1} \\
\bottomrule
\end{tabular}}
\label{tab2}
\end{table}


\textbf{Person Re-Identification.}
To further prove the effectiveness of our method on other related object re-identification task, we also compare Colo-ReID with existing methods (\textit{e.g.} CNN / Attention based architectures) in Table~\ref{tab3}. Note that we do not apply any post-processing method like Re-Rank~\citep{zhong2017re} in our approach. As we see, we can also achieve the competitive performance on Market-1501, DukeMTMC-reID and CUHK03 dataset with considerable advantages respectively. For example, our VT-ReID method can achieve a mAP/Rank-1 performance of 85.3\% and 88.3\% respectively on CUHK03 dataset, leading \textbf{+1.2\%} and \textbf{+3.6\%} improvement of mAP and Rank-1 accuracy on CUHK03 dataset when compared to the second best method C2F~\citep{zhang2021coarse} and SCSN~\citep{chen2020salience}. On the other hand, we also observe that the performance of proposed VT-ReID method is not always satisfactory in general person re-identification method (\textit{e.g.} \textcolor[rgb]{1.00,0.39,0.09}{\textbf{88.1\%}} vs. \textcolor[rgb]{0.20,0.40,0.80}{\textbf{88.6\%}} mAP on Market-1501 dataset), we will elaborate this phenomena with more details in the following Section~\ref{dis}.

\begin{table}[!t]
  \centering
  \caption{Performance comparison (\%) with other image retrieval or object re-identification methods on the Market-1501, DukeMTMC-reID and CUHK03 dataset, respectively. \textbf{Bold} indicates the best and \underline{underline} the second best.}
  \setlength{\tabcolsep}{1.8mm}{
    \begin{tabular}{lcccccc}
    \toprule
    \multirow{2}[4]{*}{Method} & \multicolumn{2}{c}{Market-1501} & \multicolumn{2}{c}{DukeMTMC-reID} & \multicolumn{2}{c}{CUHK03} \\
\cmidrule{2-7}          & mAP   & Rank-1 & mAP   & Rank-1 & mAP   & Rank-1 \\
    \midrule
    PCB~\citep{wang2020surpassing} & 81.6  & 93.8  & 69.2  & 83.3  & 57.5  & 63.7 \\
    MHN~\citep{chen2019mixed}   & 85.0  & 95.1  & 77.2  & 89.1  & 76.5  & 71.7 \\
    Pytramid~\citep{zheng2019pyramidal} & 88.2  & \underline{95.7}  & 79.0  & 89.0  & 74.8  & 78.9 \\
    BAT-net~\citep{fang2019bilinear} & 85.5  & 94.1  & 77.3  & 87.7  & 73.2  & 76.2 \\
    ISP~\citep{zhu2020identity}   & \textbf{88.6}  & 95.3  & \textbf{80.0}  & 89.6  & 71.4  & 75.2 \\
    CBDB-Net~\citep{tan2021incomplete} & 85.0  & 94.4  & 74.3  & 87.7  & 72.8  & 75.4 \\
    C2F~\citep{zhang2021coarse}   & 87.7  & 94.8  & 74.9  & 87.4  & \underline{84.1}  & 81.3 \\
    MGN~\citep{wang2018learning}   & 86.9  & \textbf{95.7}  & 78.4  & 88.7  & 66.0  & 66.8 \\
    SCSN~\citep{chen2020salience}  & \underline{88.3}  & 92.4  & 79.0  & \underline{91.0}  & 81.0  & \underline{84.7} \\
    \midrule
    VT-ReID (Ours) & 88.1  & 93.8  & \underline{79.2}  & \textbf{92.6}  & \textbf{85.3}  & \textbf{88.3}  \\
    \bottomrule
    \end{tabular}}%
  \label{tab3}%
\end{table}%

\subsection{Visualization results}
\begin{figure*}[!t]
\centering{\includegraphics[width=\linewidth]{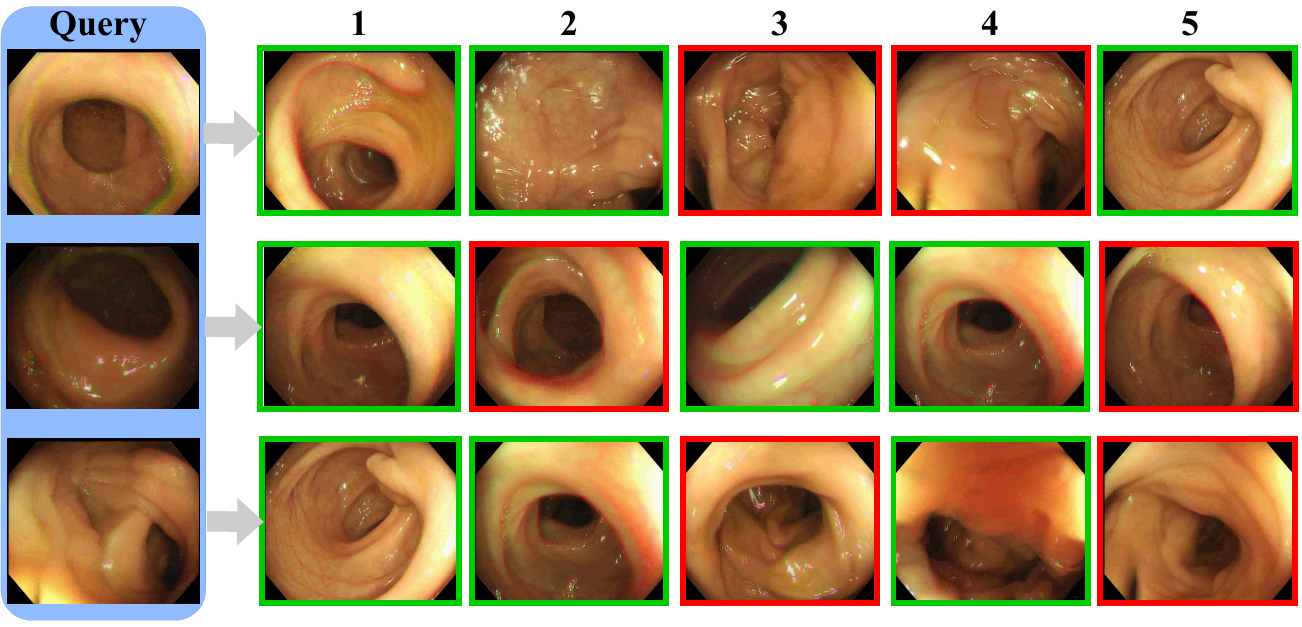}}
\caption{Top-5 ranking list for some query images on Colo-Pair dataset by VT-ReID. In each row, images are arranged in descending order according to their similarities with the query on the left. The true and false matches are in the \textcolor{green}{green} and \textcolor{red}{red} boxes, respectively.}
\label{fig3}
\end{figure*}

To go even further, we also give some qualitative results of our proposed multimodal learning method VT-ReID, Fig.~\ref{fig3} presents a visual comparison of the top-5 ranking results for some given query polyp images.
The first two results shows the great robustness regardness of the viewpoint or illumination of these captured polyps, VT-ReID features can robustly represent discriminative information of their identities. The second query image is captured in a low-resolution and poor illumination condition, losing an amount of important information. However, from some detailed clues such as the texture or position, most of the ranking results are accurate and with high quality. The last polyp shows the irregular polyp contour and obvious shadow, but we can obtain its captured images in the view in Rank-1, 3 and 4. We attribute this surprising results to the effects of local feature, which establishes the relationships when some salient parts or discriminative information are lost.

\subsection{Discussion}
\label{dis}
According to the experiment results, the proposed Colo-ReID method has shown it’s potential for polyp ReID task in small sample scenarios. To go even further, we gave an explanation about two interesting phenomenons observed during the experiments.

First, from the Table~\ref{tab2}, we can easily observed that the performance of our method is still far from the satisfactory in real-world clinical application. It is really true that our proposed VT-ReID method can effectively take advantage of the visual and text information to enhance the  discriminability of the colonoscopic polyp ReID model. However, we would like to emphasize that our method only takes a simple transformer-based architecture to extract the discriminative feature from multimodal polyp dataset. Consequently, huge performance gap between algorithm and real-world application warrants further research and consideration when deploying ReID model in real scenarios.

Second, according to the Table~\ref{tab3}, there exists an interesting phenomenon that the performance of our VT-ReID is a litter inferior to the performance of ISP~\citep{zhu2020identity} and MGN~\citep{wang2018learning} on person re-identification task (\textit{e.g.} \textcolor[rgb]{1.00,0.39,0.09}{\textbf{88.1\%}} vs. \textcolor[rgb]{0.20,0.40,0.80}{\textbf{88.6\%}} mAP; \textcolor[rgb]{1.00,0.39,0.09}{\textbf{93.8\%}} vs. \textcolor[rgb]{0.20,0.40,0.80}{\textbf{95.7\%}} Rank-1 on Market-1501
dataset). We suspect this is due to the multiple granularity network MGN and the identity-guided human semantic parsing network ISP can learn more discriminative and representative information, especially global or local representation with certain granularity of body partition at the semantic level, which can remarkably improve the performance for object ReID system.

\section{Conclusion and Future Work}
\label{sec5}
In this work, we propose a simple but effective multimodal training method named VT-ReID. Based on it, a dynamic clustering based mechanism called DCM is introduced to further boost the performance of colonoscopic polyp re-identification task, which allows our proposed method to be more robust in the general scenario at different performance efficiency trade-offs.
Comprehensive experiments conducted on newly constructed Colo-Pair datasets and other related re-identification datasets also demonstrate that Colo-ReID consistently improves over the baseline method by a large margin, proving itself as a strong baseline for colonoscopic polyp ReID task. Importantly, it also outperforms (or is on par with) other state-of-the-art methods on these datasets with the dynamic clustering regulation.
In the future, we will still focus on visual-text representation learning with self-attention mechanism for multi-domain colonoscopic polyp retrieval, and further explore the potential of proposed method to handle with other challenging image retrieval task in the medical area.

\bmhead{Acknowledgments}

This work was supported by the National Natural Science Foundation of China under Grant No. 81974276 and Startup Fund for Young Faculty at SJTU under Grant No. 23X010501967.
The authors would like to thank the anonymous reviewers for their valuable suggestions and constructive criticisms.

\section*{Declarations}

\begin{itemize}
\item \textbf{Funding} \\  This work was partially supported by the National Natural Science Foundation of China under Grant No. 81974276 and Startup Fund for Young Faculty at SJTU under Grant No. 23X010501967.
\item \textbf{Conflict of interest} \\  The authors declare that they have no conflict of interest.
\item \textbf{Ethics approval} \\  All procedures performed in studies involving human participants were in accordance with
the ethical standards of the institutional and/or national research committee.
\item \textbf{Consent to participate} \\  All human participants consented for participating in this study.
\item \textbf{Consent for publication} \\  All contents in this paper are consented for publication.
\item \textbf{Availability of data and material} \\  The data used for the experiments in this paper are available online, see Section~\ref{sec4.1} for more details.
\item \textbf{Code availability} \\  The code of this project is publicly available at \url{https://github.com/JeremyXSC/VT-ReID}.
\item \textbf{Authors' contributions} \\  Suncheng Xiang contributed conception and design of the study. Suncheng Xiang and Cang Liu contributed to experimental process and evaluated and interpreted model results. Dahong Qian and Suncheng Xiang obtained funding for the project. Dahong Qian provided clinical guidance. Suncheng Xiang drafted the manuscript. All authors contributed to manuscript revision, read and approved the submitted version.
\end{itemize}

%
%
%
%
%
%
%

\bibliography{sn-bibliography}


\end{document}